\documentclass[graybox]{svmult}

\usepackage{mathptmx}       
\usepackage{helvet}         
\usepackage{courier}        
\usepackage{type1cm}        
\usepackage{makeidx}         
\usepackage{graphicx}        
\usepackage{multicol}        
\usepackage[bottom]{footmisc}

\makeindex             

\usepackage{amsmath,amsmath,amsfonts,amssymb}
\usepackage{latexsym,epsf,graphicx}
\usepackage{dsfont,color,url,eurosym,mathrsfs}
\RequirePackage[colorlinks,citecolor=blue,urlcolor=blue]{hyperref}

\newtheorem{assumption}[theorem]{Assumption}
{\hfill\BlackBox\end{list}}
\usepackage{dsfont} 
\usepackage{mathrsfs}
\usepackage{amsmath,amsfonts,latexsym,epsf,graphicx}

\newcommand{\bc}{\begin{center}}
\newcommand{\ec}{\end{center}}
\newcommand{\bi}{\begin{itemize} \parsep0.2em \itemsep0.2em}
\newcommand{\ei}{\end{itemize}}
\newcommand{\bnum}{\begin{enumerate} \parsep0.2em \itemsep0.2em}
\newcommand{\enum}{\end{enumerate}}
\newcommand{\be}{\begin{equation}}
\newcommand{\ee}{\end{equation}}
\newcommand{\beq}{\begin{eqnarray}}
\newcommand{\eeq}{\end{eqnarray}}
\newcommand{\beqna}{\begin{eqnarray*}}
\newcommand{\eeqna}{\end{eqnarray*}}
\newcommand{\bd}{\begin{displaymath}}
\newcommand{\ed}{\end{displaymath}}
\newcommand{\bt}{\begin{tabular}}
\newcommand{\et}{\end{tabular}}

\newcommand{\Ex}{{\mathbb{E}}}

\newcommand{\Q}{{\rm Q}}

\newcommand{\Law}[2]{\mathscr{L}_{#1}(#2)}

\newcommand{\fG}{\mathscr{G}}
\newcommand{\fL}{\mathscr{L}}

\newcommand{\PM}  {\mathcal{M}_1}   

\newcommand{\inw}{{\,\rightsquigarrow\,}}

\newlength{\fixboxwidth}
\newcommand{\N}{\mathds{N}}    
\newcommand{\R}{\mathds{R}}    

\def \P {{\textrm{P}}}    
\def \Q {\textrm{Q}}    

\newcommand{\Zhat}{{\hat{Z}}}

\def \B         {{\cal B}}                
\def \O { \Omega }

\def \lb        { \lambda }

\def \ve        { \varepsilon }

\def \s         { \sigma }

\def \et        { \tilde{\eta}}

\newcommand{\snorm}[1] {\Vert #1 \Vert}

\newcommand{\inorm}[1]{\snorm{#1}_{\infty}}

\newcommand{\sLp}[1]{\mbox{${\cal L}_p(\mu)$}}

\newcommand{\cC}    {\mathcal{C}}

\newcommand{\cX}    {\mathcal{X}}  
\newcommand{\cY}    {\mathcal{Y}}  
  
\newcommand{\cZ}    {\mathcal{Z}}

\newcommand{\bbD}    {\mathbb{D}} 
\newcommand{\bbE}    {\mathbb{E}} 
\newcommand{\bbF}    {\mathbb{F}} 
\newcommand{\bbG}    {\mathbb{G}} 
\newcommand{\bbH}    {\mathbb{H}} 
\newcommand{\bbP}    {\mathbb{P}}

\newcommand{\bbY}    {\mathbb{Y}}

\newcommand{\hnorm}[1]{\left\Vert #1 \right\Vert_{H}}
\newcommand{\hhnorm}[1]{\left\Vert #1 \right\Vert_{H}^2}

\newcommand{\RLP}[1]{{{\cal R}_{L,\P}(#1)}}

\def \P           { \mathrm{P} }   
\def \Q           { \mathrm{Q} } 
\newcommand{\Pn}  {{\bbP_n}}   

\newcommand{\Dn} {\mathrm{D}_n}

\newcommand{\Gn}  {{\bbG_n}}
\newcommand{\Pnhat}  {{\hat{\bbP}_n}}  
\newcommand{\Gnhat}  {{\hat{\bbG}_n}}

\newcommand{\cXY}{\cX\times\cY}

\newcommand{\cXYR}{\cX\times\cY\times\R}

\newcommand{\BL}{{\mathrm{BL}}}

\begin{document}

\title*{On the Consistency of the Bootstrap Approach for Support Vector Machines and Related Kernel Based Methods}
\titlerunning{On the Consistency of the Bootstrap Approach for SVMs} 
\author{Andreas Christmann and Robert Hable}
\authorrunning{Andreas Christmann and Robert Hable} 
\institute{Andreas Christmann \at University of Bayreuth, Department of Mathematics, Germany, \email{andreas.christmann@uni-bayreuth.de}
\and Robert Hable \at University of Bayreuth, Department of Mathematics,
Germany, \email{robert.hable@uni-bayreuth.de}}
%
%
\maketitle

\abstract*{It is shown that bootstrap approximations of 
support vector machines (SVMs) based on a general convex and 
smooth loss function and on a general kernel are consistent.
This result is useful to approximate the unknown finite 
sample distribution of SVMs  by the bootstrap approach.}

\abstract{It is shown that bootstrap approximations of 
support vector machines (SVMs) based on a general convex and 
smooth loss function and on a general kernel are consistent.
This result is useful to approximate the unknown finite 
sample distribution of SVMs  by the bootstrap approach.}


\section{Introduction}\label{sec:1}
Support vector machines and related kernel based methods can be considered
as a hot topic in machine learning because they
have good statistical and numerical properties under weak assumptions
and have demonstrated their often good generalization properties in many
applications, 
see e.g. 
\cite{Vapnik1995,Vapnik1998},
\cite{ScSm2002}, and
\cite{SteinwartChristmann2008a}.
To our best knowledge, the original SVM approach by 
\cite{BoserGuyonVapnik1992} was derived
from the generalized portrait algorithm invented earlier by 
\cite{VapnikLerner1963}. 
Throughout the paper, the term SVM will be used in the broad sense, i.e. 
for a general convex loss function and a general kernel.

SVMs based on many standard kernels as for example the Gaussian
RBF kernel are nonparametric methods.
The finite sample distribution of many 
nonparametric methods is unfortunately unknown because the distribution $\P$ from which the 
data were generated is usually completely unknown and because there are often
only asymptotical results describing the consistency or the rate
of convergence of such methods known so far.
Furthermore, there is in general \emph{no} uniform rate of convergence
for such nonparametric methods due to the famous no-free-lunch theorem,
see \cite{Devroye1982} and \cite{DevroyeGyoerfiLugosi1996}.
Informally speaking, the no-free-lunch theorem states that, for sufficiently
malign distributions, the average risk of any statistical 
(classification) method may tend arbitrarily slowly to zero.
Theses facts are true for SVMs. SVMs are known to be universally consistent and fast rates of convergence are known for broad
\emph{subsets} of all probability distributions. 
The asymptotic normality of SVMs was shown recently by \cite{Hable2012}
under certain conditions.

Here, we apply a different approach to SVMs, namely Efron's bootstrap.
The goal of this paper is to show that bootstrap approximations 
of SVMs which are based on a general convex and smooth loss function 
and a general smooth kernel are consistent under mild assumptions;
more precisely, convergence in outer probability is shown.  
This result is useful to draw statistical decisions based on SVMs, e.g.  confidence intervals, tolerance intervals and so on. 

We mention that both the sequence of SVMs and the sequence of their
corresponding risks are qualitatively robust under mild assumptions, see
\cite{ChristmannSalibianBarreraVanAelst2013}.
Hence, Efron's bootstap approach turns out to be quite successful 
for SVMs from several aspects.

The rest of the paper has the following structure.
Section 2 gives a brief introduction into SVMs.
Section 3 gives the result.
The last section contains the proof and related results.

\section{Support Vector Machines}\label{sec:2}
Current statistical applications are characterized by a wealth of large
and high-dimensional data sets.
In classification and in regression problems there is a variable of main interest, 
often called ``output values'' or ``response'', 
and a number of potential explanatory variables, which are often 
called ``input values''.
These input values are used to model the observed output values or to predict future output values.  
The observations consist of $n$ pairs $(x_1, y_1)$, \ldots, $(x_n, y_n)$,
which will be assumed to be independent realizations of a random pair $(X,Y)$.  
We are interested in
minimizing the risk or to obtain a function $f: {\cal X} \to {\cal Y}$ such that $f(x)$ 
is a good predictor for the response $y$, if $X=x$ is observed.  
The prediction should be made in an automatic way. 
We refer to this process of determining a prediction method as ``statistical machine learning'', 
see e.g. \cite{Vapnik1995, Vapnik1998, ScSm2002, CuckerZhou2007, SmaleZhou2007}. 
Here, by ``good predictor'' we mean that $f$ minimizes the expected loss, i.e.  the risk, 
$$
\RLP{f} =
\Ex_\P \left[ L \left( X, Y, f(X) \right) \right],
$$
where $\P$ denotes the unknown joint
distribution of the random pair $(X,Y)$ and $L : {\cal X} \times {\cal Y}
\times \R \to [0, +\infty)$ is a fixed loss function. 
As a simple example, the least squares loss $L(X, Y, f(X)) = (Y - f(X))^2$ yields the optimal predictor $f(x) = \Ex_\P(Y|X=x)$, $x\in{\cal X}$. 
Because $\P$ is unknown, we can neither compute
nor minimize the risk $\RLP{f}$ directly.

Support vector machines, see 
\cite{VapnikLerner1963}, \cite{BoserGuyonVapnik1992},
\cite{Vapnik1995, Vapnik1998},
provide a highly versatile framework to perform
statistical machine learning in a wide variety of setups. The minimization of regularized empirical risks over reproducing kernel Hilbert spaces was already considered e.g. by \cite{PoggioGirosi1990}.
Given a kernel $k:
{\cal X} \times {\cal X} \to \R$ we consider predictors $f \in H$, where $H$
denotes the corresponding reproducing kernel Hilbert space of functions from 
$\cX$ to $\R$.  The
space $H$ includes, for example, all functions of the form $f(x) = \sum_{j=1}^m
\alpha_j \, k(x, x_j)$ where $x_j$ are arbitrary elements in ${\cal X}$ and
$\alpha_j \in \R$, $1 \le j \le m$.  To avoid overfitting, a support vector
machine $f_{L,\P,\lb}$ is  defined as the solution of a regularized risk minimization problem. More precisely,
\begin{equation} \label{pop.min}
f_{L,\P,\lb} \, = \, \arg \inf_{f \in H} \ \Ex_{\P}  L \left( X, Y, f(X) \right) \, + \, 
\lb \, \| f \|^2_H \, ,
\end{equation}
where $\lb\in(0,\infty)$ is the regularization parameter.
For a sample $D= ((x_1, y_1), \ldots,$ $(x_n, y_n))$ the corresponding estimated function is given by
\begin{equation} \label{sample.min}
f_{L,\Dn,\lb}  \, = \, \arg \inf_{f \in H} \ \frac{1}{n} \sum_{i=1}^n L \left( x_i, y_i, f(x_i) \right)  \, + \, 
\lambda \, \| f \|^2_H \, ,
\end{equation}
where $\Dn$ denotes the empirical distribution based on $D$ (see~\eqref{emp_measure} below).
Note that the optimization problem \eqref{sample.min} corresponds to \eqref{pop.min} when using $\Dn$  instead of $\P$. 

Efficient algorithms to compute $\hat{f}_n:=f_{L,\Dn,\lb}$ exist for a number of different loss functions.  
However, there are often good reasons to consider other convex loss functions, e.g. the hinge loss
$L(X,Y,f(X))=\max\{1-Y\cdot f(X),0\}$ for binary classification purposes or 
the $\epsilon$-insensive loss $L(X,Y,f(X))=\max\{0, |Y-f(X)|-\epsilon\}$ for regression purposes, where $\epsilon>0$.
As these loss functions are not differentiable, the logistic loss functions
$L(X,Y,f(X))=\ln(1+\exp(-Y\cdot f(X)))$ and $L(X,Y,f(X))=-\ln(4 e^{Y-f(X)}/(1+e^{Y-f(X)})^2)$ and Huber-type loss functions are also 
used in practice. These loss functions can be
considered as smoothed versions of the previous two loss functions.

An important component of statistical analyses concerns quantifying and
incorporating uncertainty (e.g. sampling variability) in the reported
estimates. For example, one may want to include confidence bounds along the
individual predicted values $\hat{f}_n(x_i)$ obtained from \eqref{sample.min}.
Unfortunately, the sampling distribution of the estimated function $\hat{f}_n$ is unknown. 
Recently, \cite{Hable2012} derived the asymptotic distribution of SVMs under some mild conditions.
Asymptotic confidence intervals based on those general results are always symmetric.

Here, we are interested in approximating the finite sample distribution of SVMs
by Efron's bootstrap approach, because confidence intervals based on the bootstrap approach can be asymmetric.
The bootstrap \cite{Efron1979} provides an alternative way to estimate the
sampling distribution of a wide variety of estimators.  To fix ideas, consider a
functional $S : {\cal M} \to {\cal W}$, where ${\cal M}$ is a set of
probability measures and ${\cal W}$ denotes a metric space.  Many estimators
can be included in this framework.  Simple examples include the sample mean
(with functional $S(\P) = \int Z \, d\P$) and M-estimators (with functional
defined implicitly as the solution to the equation $\Ex_\P \Psi(Z, S(\P)) = 0$).
Let $\B(\cZ)$ be the Borel $\s$-algebra on $\cZ = \cX \times \cY$  and denote 
the set of all Borel probability measures on
$(\cZ,\B(\cZ))$ by $\PM(\cZ,\B(\cZ))$.
Then, it follows that \eqref{pop.min} defines an operator 
$$
S: \PM(\cZ,\B(\cZ)) \to H, \qquad S(P) = f_{L,\P,\lb},
$$
i.e. the support vector machine.  Moreover, the estimator in \eqref{sample.min} satisfies 
$$
f_{L,\Dn,\lb} = S(\Dn)
$$
where 
\begin{equation}
\Dn=\frac{1}{n}\sum_{i=1}^n \delta_{(x_i,y_i)} 
\label{emp_measure}
\end{equation}
is the empirical distribution based on the sample 
$D=((x_1, y_1), \ldots, (x_n,y_n))$ and $\delta_{(x_i,y_i)}$ denotes the Dirac measure 
at the point $(x_i,y_i)$.  

More generally, let $Z_i=(X_i,Y_i)$, $i=1,\ldots,n$, be
independent and identically distributed (i.i.d.) random variables with
distribution $\P$, and let 
$$
S_n(Z_1, \ldots, Z_n) = S(\P_n)
$$ 
be the corresponding
estimator, where 
$$
\P_n=\frac{1}{n}\sum_{i=1}^n \delta_{Z_i}.
$$
Denote the distribution of $S(\P_n)$ by $\Law{n}{S; \P} = \Law{}{S(\P_n)}$.  If $\P$
was known to us, we could estimate this sampling distribution by drawing a
large number of random samples from $\P$ and evaluating our estimator on them.
The basic idea of Efron's bootstrap approach is to replace the unknown distribution $\P$
by an estimate $\hat{\P}$. Here we will consider the natural non-parametric
estimator given by the  sample empirical distribution $\P_n$. In other words, we
estimate the distribution of our estimator of interest by its sampling
distribution when the data are generated by $\P_n$.  In symbols, the bootstrap
proposes to use 
$$
  \widehat{ {\cal L}_n (S; \P)} = \Law{n}{S; \P_n}.
$$
Since this
distribution is generally unknown, in practice one uses Monte Carlo simulation
to estimate it by repeatedly evaluating the estimator on samples drawn from
$\Dn$. 
Note that drawing a sample from $\Dn$ means that $n$ observations 
are drawn \emph{with replacement} from the original $n$ observations $(x_1, y_1)$, \ldots, $(x_n, y_n)$.


\section{Consistency of Bootstrap SVMs}\label{sec:consist}
In this section it will be shown under appropriate assumptions that the weak consistency of bootstrap estimators carries over to the Hada\-mard-differentiable SVM functional in the sense that the sequence of ``conditional random laws'' (given $(X_1,Y_1), (X_2,Y_2),\ldots$) of
$\sqrt{n}(f_{L,\Pnhat,\lb} - f_{L,\Pn,\lb})$ is asymptotically consistent in probability for estimating the laws of the random elements 
$\sqrt{n}(f_{L,\Pn,\lb} - f_{L,\P,\lb})$.
In other words, if $n$ is large, the ''random distribution''
\be 
\fL(\sqrt{n}(f_{L,\Pnhat,\lb} - f_{L,\Pn,\lb}))
\ee
based on bootstrapping an SVM can be considered as a valid approximation of the unknown finite sample distribution
\be 
\fL(\sqrt{n}(f_{L,\Pn,\lb} - f_{L,\P,\lb})).
\ee

\begin{assumption}\label{AC.assump.consist}
Let $\cX\subset\R^d$ be closed and bounded and let $\cY \subset \R$ be closed. Assume
that $k:\cX\times\cX \to \R$ is the restriction of an $m$-times continuously differentiable kernel $\tilde{k}:\R^d\times\R^d\to\R$ such that 
$m>d/2$ and $k\ne 0$. Let $H$ be the RKHS of $k$ and let $\P$ be a probability
distribution on $(\cXY,\B(\cXY))$. Let $L:\cXYR\to[0,\infty)$ be a convex, 
$\P$-square-integrable Nemitski loss function of order $p\in[1,\infty)$ such that the partial derivatives
$$
L'(x,y,t):=\frac{\partial L}{\partial t}(x,y,t)
\qquad {\mbox and} \qquad
L''(x,y,t):=\frac{\partial^2 L}{\partial^2 t}(x,y,t)
$$
exist for every $(x,y,t)\in\cXYR$. Assume that the maps
$$
(x,y,t) \mapsto L'(x,y,t) 
\qquad {\mbox and} \qquad
(x,y,t) \mapsto L''(x,y,t)
$$
are continuous. Furthermore, assume that for every $a\in(0,\infty)$, there is a 
$b'_a\in L_2(\P)$ and a constant $b''_a\in[0,\infty)$ such that, for every $(x,y)\in\cXY$, 
\be \label{Hable2012.5}
\sup_{t\in[-a,a]} |L'(x,y,t)| \le b'_a(x,y)
\qquad {\mbox and} \qquad
\sup_{t\in[-a,a]} |L''(x,y,t)| \le b''\,.
\ee 
\end{assumption}

The conditions on the kernel $k$ in Assumption \ref{AC.assump.consist} are satisfied for many common kernels, e.g., Gaussian RBF kernel, exponential kernel, polynomial kernel, and linear kernel, but also Wendland kernels $k_{d,\ell}$ based on certain univariate polynomials
$p_{d,\ell}$ of degree $\lfloor d/2\rfloor +3\ell+1$ for
$\ell\in\N$ such that $\ell > d/4$, see \cite{Wendland2005}.

The conditions on the loss function $L$ in Assumption \ref{AC.assump.consist} are satisfied, e.g., for the logistic loss for classification or for regression, however the popular non-smooth loss functions hinge, $\ve$-insensitive, and pinball are not
covered. However, \cite[Remark 3.5]{Hable2012} described an analytical method 
to approximate such non-smooth loss functions up to an arbitrarily good precision $\epsilon>0$ by a convex $\P$-square integrable Nemitski loss function of order $p\in[1,\infty)$.

We can now state our result on the consistency of the bootstrap 
approach for SVMs. \newpage

\begin{theorem}\label{AC.BootstrapConsist}
Let Assumption \ref{AC.assump.consist} be satisfied. 
Let $\lb\in(0,\infty)$. Then
\begin{eqnarray}
 \label{AC.Bootconsist1a}
 \sup_{h\in\BL_1(H)} \bigl| 
 \Ex_M h\bigl(\sqrt{n}(f_{L,\Pnhat,\lb} - f_{L,\Pn,\lb})\bigr) 
 - \Ex h(S'_\P(\bbG))\bigr| \to 0, ~~~\\
 \label{AC.Bootconsist1b} 
 \Ex_M h\bigl(\sqrt{n}(f_{L,\Pnhat,\lb} \!-\! f_{L,\Pn,\lb})\bigr)^* 
 \!\!-\!\! \Ex_M h\bigl(\sqrt{n}(f_{L,\Pnhat,\lb} \!-\! f_{L,\Pn,\lb})\bigr)_* \to 0,~~~
\end{eqnarray}
converge in outer probability, where $\bbG$ is a tight Borel-measurable Gaussian process, $S'_\P$ is a continuous linear operator with
\be 
S'_\P(\Q) = -K_\P^{-1}\bigl( \Ex_\Q \bigl( L'(X,Y,f_{L,\P,\lb}(X))\Phi(X)\bigr) \bigr), \quad \Q\in\PM(\cXY)
\ee
and
\be 
K_\P: H \to H, \quad 
f \mapsto 2\lb f + \Ex_\P\bigl( L''(X,Y,f_{L,\P,\lb}(X)) f(X) \Phi(X)\bigr) 
\ee
is a continuous linear operator which is invertible.
\end{theorem}
For details on $K_\P$, $S'_\P$, and $\bbG$ we refer to
Lemma \ref{Hable.LemmaA.5}, Theorem \ref{Hable2012.ThmA.8}, and
Lemma \ref{Hable2012.LemmaA.9}.


\section{Proofs}\label{sec:proofs}
\subsection{Tools for the proof of Theorem \ref{AC.BootstrapConsist}}\label{Proof.sec:AC.BootstrapConsist}

We will need two general results on bootstrap methods proven in \cite{VandervaartWeller1996} and adopt their notation, see
\cite[Chapters 3.6 and 3.9]{VandervaartWeller1996}. 
Let $\Pn$ be the empirical measure of an i.i.d. sample $Z_1,\ldots Z_n$ from a probability distribution $\P$. The \emph{empirical process} is the signed measure
$$ 
  \Gn=\sqrt{n}(\Pn-\P).
$$
Given the sample values, let $\Zhat_1,\ldots,\Zhat_n$
be an i.i.d. sample from $\Pnhat$. The \emph{bootstrap empirical distribution} is the
empirical measure 
$\Pnhat:=n^{-1} \sum_{i=1}^n \delta_{\Zhat_i}$, 
and the \emph{bootstrap empirical process} is
\be \nonumber
\Gnhat = \sqrt{n}(\Pnhat-\Pn) 
= \frac{1}{\sqrt{n}}\sum_{i=1}^n (M_{ni}-1)\delta_{Z_i}\,,
\ee
where $M_{ni}$ is the number of times that $Z_i$ is ``redrawn'' from the original sample $Z_1,\ldots Z_n$, $M:=(M_{n1},\ldots,M_{nn})$ is stochastically independent of 
$Z_1,\ldots,Z_n$ and multinomially distributed with parameters $n$ and probabilities $\frac{1}{n},\ldots,\frac{1}{n}$. 
If outer expectations are computed, stochastic independence is understood 
in terms of a product probability space. Let $Z_1, Z_2, \ldots$ be the coordinate projections on the first $\infty$ coordinates of the product
space $(\cZ^\infty, \B(\cZ),\P^\infty) \times (\widetilde{\cZ},\cC,\Q)$ and let the multinomial vectors $M$ depend on the last factor only, see \cite[p.\,345f]{VandervaartWeller1996}.

The following theorem shows (conditional) weak convergence for the 
empirical bootstrap, where the symbol $\rightsquigarrow$ denotes the 
weak convergence of finite measures. 
We will need only the equivalence between $(i)$ and $(iii)$ 
from this theorem and list part $(ii)$ only for the sake of completeness.

\begin{theorem}[{\cite[Thm. 3.6.2, p.\,347]{VandervaartWeller1996}}] \label{vdVW1996.Thm3.6.2}
Let $\mathcal{F}$ be a class of measurable functions with finite envelope function.
Define $\bbY_n:=n^{-1/2}\sum_{i=1}^n (M_{N_n, i}-1)(\delta_{Z_i}-\P)$.
The following statements are equivalent:
\bnum
\item[(i)] $\mathcal{F}$ is Donsker and $\P^{*}\snorm{f-\P f}_{\mathcal{F}}^2 < \infty$;
\item[(ii)] $\sup_{h\in \BL_1} \bigl| \Ex_{M,N} h(\hat{\bbY}_n) - \Ex h(\bbG)\bigr|$ converges outer almost surely to zero
and the sequence $\Ex_{M,N} h(\hat{\bbY}_n)^* - \Ex_{M,N} h(\hat{\bbY}_n)_*$ converges almost surely to zero for every $h\in \BL_1$.
\item[(iii)] $\sup_{h\in \BL_1} \bigl| \Ex_M h(\Gnhat) - \Ex h(\bbG)\bigr|$ converges outer almost surely to zero
and the sequence $\Ex_M h(\Gnhat)^* - \Ex_M h(\Gnhat)_*$ converges almost surely to zero for every $h\in \BL_1$.
\enum
Here the asterisks denote the measurable cover functions with respect to $M$, $N$, and
$Z_1, Z_2, \ldots$ jointly.
\end{theorem}

Consider sequences of random elements $\Pn=\Pn(Z_n)$ and 
$\Pnhat=\Pnhat(Z_n,M_n)$ in a normed space $\bbD$ such that the sequence 
$\sqrt{n}(\Pn-\P)$ converges unconditionally and the sequence
$\sqrt{n}(\Pnhat-\Pn)$ converges conditionally on $Z_n$ in distribution to a tight
random element $\bbG$. A precise formulation of the second assumption is
\begin{eqnarray}
 \label{vdVW3.9.9a}
 \sup_{h\in\BL_1(\bbD)} \bigl| \Ex_M h(\sqrt{n}(\Pnhat-\Pn)) 
 - \Ex h(\bbG)\bigr| \to 0,\\
 \label{vdVW3.9.9b} 
 \Ex_M h\bigl(\sqrt{n}(\Pnhat-\Pn)\bigr)^* 
 - \Ex_M h\bigl(\sqrt{n}(\Pnhat-\Pn)\bigr)_* \to 0,
\end{eqnarray}
in outer probability, with $h$ ranging over the bounded Lipschitz functions,
see \cite[p.\,378, Formula (3.9.9)]{VandervaartWeller1996}.
The next theorem shows that under appropriate assumptions, weak consistency of the bootstrap estimators carries over to any Hadamard-differentiable functional in the sense that the sequence of ``conditional random laws'' (given $Z_1, Z_2,\ldots$) of
$\sqrt{n}(\phi(\Pnhat)-\phi(\Pn))$ is asymptotically consistent in 
probability for estimating the laws of the random elements $\sqrt{n}(\phi(\Pn)-\phi(\P))$,
see \cite[p.378]{VandervaartWeller1996}.

\begin{theorem}[{\cite[Thm. 3.9.11, p.\,378]{VandervaartWeller1996}}] \label{vdVW1996.Thm3.9.11}
(Delta-method for bootstrap in probability)
Let $\bbD$ and $\bbE$ be normed spaces. Let $\phi:\bbD_\phi \subset \bbD \to \bbE$
be Hadamard-differentiable at $\P$ tangentially to a subspace $\bbD_0$.
Let $\Pn$ and $\Pnhat$ be maps as indicated previously with values in $\bbD_\phi$ such 
that $\Gn:=\sqrt{n}(\Pn-\P)\inw \bbG$ and that {(\ref{vdVW3.9.9a})}-{(\ref{vdVW3.9.9b})} holds
in outer probability, where $\bbG$ is separable and takes its values in $\bbD_0$.
Then
\begin{eqnarray}
 \label{vdVW3.9.10a}
 \sup_{h\in\BL_1(\bbE)} \bigl| \Ex_M h\bigl(\sqrt{n}(\phi(\Pnhat)-\phi(\Pn))\bigr) 
 - \Ex h(\phi'_\P(\bbG))\bigr| \to 0,\\
 \label{vdVW3.9.10b} 
 \Ex_M h\bigl(\sqrt{n}(\phi(\Pnhat)-\phi(\Pn))\bigr)^* 
 - \Ex_M h\bigl(\sqrt{n}(\phi(\Pnhat)-\phi(\Pn))\bigr)_* \to 0,
\end{eqnarray}
holds in outer probability.
\end{theorem}
As was pointed out by \cite[p.\,378]{VandervaartWeller1996}, consistency in probability appears to be sufficient for (many) statistical purposes and the theorem above shows this is retained under Hadamard differentiability at the single distribution $\P$.

We now list some results from \cite{Hable2012}, which will also be essential for the proof of Theorem \ref{AC.BootstrapConsist}.

\begin{theorem}[{\cite[Theorem 3.1]{Hable2012}}]\label{Hable.Thm3.1}
Let Assumption \ref{AC.assump.consist} be satisfied.
Then, for every regularizing parameter $\lb_0\in(0,\infty)$, there is a tight, Borel-measurable Gaussian process
$\bbH: \O \to H$, $\omega \to \bbH(\omega)$, such that
\be \label{Hable2012.6}
\sqrt{n} \bigl( f_{L,\mathbf{D}_n,\lb_{\mathbf{D}_n}} - 
                f_{L,\P,\lb_0}\bigr)  \inw \bbH \quad \mbox{~in~} H 
\ee
for every Borel-measurable sequence of random regularization parameters 
$\lb_{\mathbf{D}_n}$  with $\sqrt{n}\bigl(\lb_{\mathbf{D}_n}-\lb_0\bigr) \to 0$ in probability.
The Gaussian process $\bbH$ is zero-mean; i.e., 
$\Ex \langle f, \bbH \rangle_H=0$ for every $f\in H$.
\end{theorem}

\begin{lemma}[{\cite[Lemma A.5]{Hable2012}}]\label{Hable.LemmaA.5}
For every $F\in B_S$ defined later in {(\ref{DefB_S})}, 
\be
K_F : H \to H, \quad 
f \mapsto 2\lb_0 f + \int L''(x,y,f_{L,\iota(F),\lb_0}(x)) f(x) \Phi(x) 
d\iota(F)(x,y)
\ee
is a continuous linear operator which is invertible.
\end{lemma}

\begin{theorem}[{\cite[Theorem A.8]{Hable2012}}]\label{Hable2012.ThmA.8}
For every $F_0\in B_S$ which fulfills $F_0(b) < \Ex_\P(b) + \lb_0$, the map
$S: B_S\to H$, $F\mapsto f_{\iota(F)}$, is Hadamard-differentiable in $F_0$ tangentially to the closed linear span $B_0=\mathrm{cl}(\mathrm{lin}(B_S))$.
The derivative in $F_0$ is a continuous linear operator $S'_{F_0}: B_0 \to H$
such that
\be
S'_{F_0}(G) = -K_{F_0}^{-1}\bigl( \Ex_{\iota(G)} ( L'(X,Y,f_{L,\iota(F_0),\lb_0}(X))\Phi(X))\bigr), ~~~\forall\,G \in \mathrm{lin}(B_S).
\ee
\end{theorem}

\begin{lemma}[{\cite[Lemma A.9]{Hable2012}}]\label{Hable2012.LemmaA.9}
For every data set $D_n=((x_1,y_1),\ldots,$ $(x_n,y_n))\in (\cXY)^n$, let
$\bbF_{D_n}$ denote the element of $\ell_\infty(\fG)$ which corresponds to the empirical measure $\Pn:=\bbP_{D_n}$. That is, 
$\bbF_{D_n}(g)=\int g\,d\Pn=n^{-1}\sum_{i=1}^n g(x_i,y_i)$ for every $g\in\fG$.
Then 
\be 
\sqrt{n} \bigl( \bbF_{\Dn} - \iota^{-1}(\P)\bigr) \inw \bbG
\quad \mbox{in~} \ell_\infty(\fG),
\ee
where $\bbG:\Omega\to \ell_\infty(\fG)$ is a tight Borel-measurable Gaussian process such that $\bbG(\omega)\in B_0$ for every $\omega \in \Omega$.
\end{lemma}

\subsection{Proof of Theorem \ref{AC.BootstrapConsist}}

The proof relies on the application of Theorem \ref{vdVW1996.Thm3.9.11}. Hence, we have to show the following steps:
\bnum
\item[1.~] The empirical process $\Gn=\sqrt{n}(\Pn-\P)$ weakly converges to a separable Gaussian process $\bbG$. 
\item[2.~] SVMs are based on a map $\phi$ which is Hadamard differentiable at $\P$ tangentially to some appropriate subspace.
\item[3.~] The assumptions {(\ref{vdVW3.9.9a})}-{(\ref{vdVW3.9.9b})} of Theorem \ref{vdVW1996.Thm3.9.11} are satisfied. For this purpose we will use Theorem \ref{vdVW1996.Thm3.6.2}. Actually, we will show that part \emph{(i)} of Theorem \ref{vdVW1996.Thm3.6.2} is satisfied which gives the equivalence to part \emph{(iii)}, from which we conclude that {(\ref{vdVW3.9.9a})}-{(\ref{vdVW3.9.9b})} hold true.
For the proof that part \emph{(i)} of Theorem \ref{vdVW1996.Thm3.6.2} is satisfied, i.e., that a suitable set $\mathcal{F}$ is a $\P$-Donsker class and that $\P^{*}\snorm{f-\P f}_{\mathcal{F}}^2 < \infty$, we use several facts recently shown by \cite{Hable2012}.
\item[4.~] We put all parts together and apply Theorem \ref{vdVW1996.Thm3.9.11}.
\enum

\emph{Step 1.~} 
To apply Theorem \ref{vdVW1996.Thm3.9.11}, we first have to specify the considered spaces $\bbD$, $\bbE$, $\bbD_\phi$, $\bbD_0$ and the map $\phi$. 
As in \cite{Hable2012} we use the following notations.
Because $L$ is a $\P$-square-integrable Nemitski loss function of order $p\in[1,\infty)$, there is a function $b\in L_2(\P)$ such that
\be
|L(x,y,t)|\le b(x,y) + |t|^p\, , \qquad (x,y,t)\in \cXYR.
\ee
Let 
\be \label{Hable2012.c0}
c_0 := \sqrt{\lb_0^{-1} \Ex_\P(b)} +1,
\ee
Define
\be
\fG:=\fG_1 \cup \fG_2 \cup \fG_3 \,,
\ee
where
\be
\fG_1:= \{ g: \cXY \to \R: ~\exists\, z\in \R^{d+1} \mbox{~such~that~} g=I_{(-\infty,z]}\}
\ee
is the set of all indicator functions $I_{(-\infty,z]}$,
\be
\fG_2:= \left\{ g: \cXY \to \R \left| 
   \begin{array}{l}
    \exists\, f_0\in H, \exists\, f\in H \mbox{~such~that~} \hnorm{f_0} \le c_0, \\
    \hnorm{f}\le 1, g(x,y)=L'(x,y,f_0(x))f(x) ~\forall\, (x,y)
   \end{array}
   \right.
\right\},
\ee
and
\be
\fG_3:=\{b\}.
\ee
Now let $\ell_\infty(\fG)$ be the set of all bounded functions $F: \fG \to \R$ with
norm $\inorm{F} = \sup_{g\in\fG} |F(g)|$.
Define 
\be \label{DefB_S}
B_S :=
\left\{
F: \fG\to \R 
\left| 
\begin{array}{l}
 \exists\, \mu \ne 0 \mbox{~a~finite~measure~on~} \cXY \mbox{~such~that~}\\
   F(g)=\int g \,d\mu~\forall\, g\in\fG,\\
   b\in L_2(\mu), b'_a\in L_2(\mu) ~\forall\,a\in(0,\infty)
\end{array}
\right.
\right\}
\ee
and
\be
B_0 := \mathrm{cl}(\mathrm{lin}(B_S))
\ee
the closed linear span of $B_S$ in $\ell_\infty(\fG)$. That is, $B_S$ is a subset of $\ell_\infty(\fG)$ whose elements correspond to finite measures. Hence probability measures are covered as special cases. The elements of $B_S$ can be interpreted as some kind of generalized distributions functions, because $\fG_1\subset\fG$. The assumptions on $L$ and $\P$ imply that $\fG\to\R$, 
$g\mapsto \int g\,d\P$ is a well-defined element of $B_S$. For every $F\in B_S$, let
$\iota(F)$ denote the corresponding finite measure on $(\cXY,\B(\cXY))$ such that
$F(g)=\int g\,d\mu$ for all $g\in\fG$. Note that the map $\iota$ is well-defined, because by definition of $B_S$, $\iota(F)$ uniquely exists for every $F\in B_S$.

With these notations, we will apply Theorem \ref{vdVW1996.Thm3.9.11}
for
\be
\begin{array}{ll}
\bbD:=\ell_\infty(\fG), & \bbE:=H ~(\mbox{=~RKHS~of~the~kernel~}k), \,\\
\bbD_\phi := B_S, & \bbD_0:=B_0:=\mathrm{cl}(\mathrm{lin}(B_S)),\\
\lb_0\in(0,\infty),\\
\phi:= S, & S: B_S\to H, ~F\mapsto f_{\iota(F)}:=f_{L,\iota(F),\lb_0} :=\\
&  ~~~\arg\inf_{f\in H} \int L(x,y,f(x)) \,d\iota(F)(x,y) + \lb_0\hhnorm{f}\,. 
\end{array}
\ee

At first glance this definition of $S$ seems to be somewhat technical.
However, this will allow us to use a functional delta method for bootstrap estimators of SVMs with regularization parameter $\lb=\lb_0\in(0,\infty)$.

Lemma \ref{Hable2012.LemmaA.9} guarantees that  
the empirical process $\Gn:=\sqrt{n}(\Pn-\P)$ weakly converges to a tight Borel-measurable Gaussian process.

Since a $\s$-compact set in a metric space is separable, separability 
of a random variable is slightly weaker than tightness, see \cite[p.\,17]{VandervaartWeller1996}. 
Therefore, $\bbG$ in our Theorem \ref{AC.BootstrapConsist} is indeed separable.\\

\emph{Step 2.~} 
Theorem \ref{Hable2012.ThmA.8} showed that the map $S$ indeed satisfies the necessary Hadamard-differentiability in the point $\P:=\iota^{-1}(F)$.

\emph{Step 3.~} 
We know that $\fG$ is a $\P$-Donsker class, see Lemma \ref{Hable2012.LemmaA.9}. Hence, an immediate consequence from 
\cite[Theorem 3.6.1, p.\,347]{VandervaartWeller1996} is, that 
\be \label{term3.9.9a}
\sup_{h\in \BL_1} | \Ex_M h(\Gnhat) - \Ex h(\bbG)|
\ee
converges in outer probability to zero and $\Gnhat$ is asymptotically measurable.

However, we will prove a somewhat stronger result, namely that $\fG$ is a $\P$-Donsker class and $\P^*\snorm{g-\P g}_{\fG}^2 <\infty$, which is part \emph{(i)} of 
Theorem \ref{vdVW1996.Thm3.6.2}, and then part \emph{(iii)} of Theorem \ref{vdVW1996.Thm3.6.2} yields, that the term in {(\ref{term3.9.9a})} converges even outer almost surely to zero and the sequence 
\be \label{termiiivdVWpage347}
  \Ex_M h(\Gnhat)^* - \Ex_M h(\Gnhat)_*
\ee
converges almost surely to zero for every $h\in \BL_1$.

Because $\fG$ is a $\P$-Donsker class, it remains to show that $\P^*\snorm{g-\P g}_{\fG}^2 <\infty$.
Due to 
\be
\P^*\snorm{g-\P g}_{\fG}^2 
:=
\int ( \sup_{g\in\fG} |g-\Ex_\P(g)| )^2 \,d\P^*
\ee
and $\fG=\fG_1 \cup \fG_2 \cup \fG_3$, we obtain the inequality
\begin{eqnarray}
\P^*\snorm{g-\P g}_{\fG}^2 & \le & 
  \P^* \sup_{g\in\fG} \bigl( g^2 + 2|g| \cdot \P |g| 
     + (\P |g|)^2 \bigr)  \nonumber \\
& \le & \P^* \sup_{g\in\fG} g^2 
         + 2 \, \P^* \sup_{g\in\fG} ( |g| \cdot \P \, |g| )
         + \sup_{g\in\fG} (\P |g|)^2 \qquad  \nonumber\\
& \le & \sum_{j=1}^3 \Bigl( \P^* \sup_{g\in\fG_j} g^2 
         + 2 \, \P^* \sup_{g\in\fG_j} ( |g|\cdot \P\,  |g| )
         + \sup_{g\in\fG_j} (\P |g|)^2 \Bigr). \label{tmptmp3} \qquad~~ 
\end{eqnarray}
We will show that each of the three summands on the right hand side of the last inequality is finite.
If $g\in\fG_1$, then $g$ equals the indicator function $I_{(-\infty,z]}$ for some
$z\in\R^{d+1}$. Hence, $\inorm{g}=1$ and the summand for $j=1$ is finite.
If $g\in\fG_3$, then $g=b\in L_2(\P)$ because $L$ is by assumption a $\P$-square-integrable Nemitski loss function of order $p\in[1,\infty)$. Hence the summand for $j=3$ is finite, too.
Let us now consider the case that $g\in\fG_2$. By definition of $\fG_2$, for every $g\in \fG_2$ there exist $f, f_0 \in H$ such that $\hnorm{f_0}\le c_0$, $\hnorm{f}\le 1$, and
$g=L'_{f_0}f$, where we used the notation $\bigl(L'_{f_0}f\bigr)(x,y):=L'(x,y,f_0(x))f(x)$ for all 
$(x,y)\in\cXY$. Using $\inorm{f}\le \inorm{k}\hnorm{f}$ for every $f\in H$, we obtain
\be \label{tmptmp1}
\hnorm{f_0}\le c_0 ~\Rightarrow~ \inorm{f_0}\le c_0\inorm{k}
\quad \mbox{and} \quad 
\hnorm{f}\le 1 ~\Rightarrow~ \inorm{f}\le \inorm{k} \,.
\ee
Define the constant  $a:=c_0 \inorm{k}$ with $c_0$ given by {(\ref{Hable2012.c0})}.
Hence, for all $(x,y)\in\cXY$,  
\begin{eqnarray}
\sup_{f_0\in H;\, \hnorm{f_0}\le c_0} |L'(x,y,f_0(x))|^2 & \le & 
\sup_{f_0\in H;\, \inorm{f_0}\le a} \,\, \sup_{t\in[-a,+a]} |L'(x,y,t)|^2  \nonumber\\
& \stackrel{\footnotesize{{(\ref{Hable2012.5})}}}{\le} & \sup_{f_0\in H; \,\inorm{f_0}\le a} (b'_a(x,y))^2 \,. \label{tmptmp2}
\end{eqnarray}
Hence we get
\begin{eqnarray}
 & & \P^* \sup_{g\in\fG_2} g^2 \nonumber \\
 & = & \int \sup_{g\in\fG_2; \, \hnorm{f_0}\le c_0, \hnorm{f}\le 1, g=L'_{f_0}f}
  |L'(x,y,f_0(x)) f(x)|^2 \,d\P^*(x,y)   \qquad \nonumber\\
  & \le & \int \sup_{f_0\in H; \,\hnorm{f_0}\le c_0}  |L'(x,y,f_0(x))|^2 
  \sup_{f\in H; \, \hnorm{f}\le 1} |f(x)|^2 \,d\P^*(x,y) \qquad \nonumber \\
  & \stackrel{\footnotesize{(\ref{tmptmp2}),(\ref{tmptmp1})}}{\le} & \inorm{k}^2 \int (b'_a)^2 \,d\P^* 
           = \inorm{k}^2 \int (b'_a)^2 \,d\P < \infty \,,
  \nonumber
\end{eqnarray}
because $b'_a\in L_2(\P)$ and $\inorm{k} < \infty$ by Assumption \ref{AC.assump.consist}.
With the same arguments we obtain, for every $g\in\fG_2$, 
\begin{eqnarray}
\P |g| & \le & \int \sup_{g\in\fG_2} |g|\,d\P^* \nonumber \\
& \le &  \int \sup_{f_0\in H; \,\hnorm{f_0}\le c_0}  |L'(x,y,f_0(x))| 
  \sup_{f\in H; \,\hnorm{f}\le 1} |f(x)| \,d\P^*(x,y) \nonumber \\
  & \stackrel{\footnotesize{(\ref{tmptmp2}),(\ref{tmptmp1})}}{\le} & \int b'_a(x,y) \, \inorm{k} \,d\P^*(x,y) \nonumber\\
  & \le & \inorm{k} \int b'_a \,d\P < \infty \,, \nonumber
\end{eqnarray}
because $b'_a\in L_2(\P)$ and $\inorm{k}< \infty$ by Assumption \ref{AC.assump.consist}.
Hence, 
\begin{eqnarray*}
\P^* \sup_{g\in\fG_2} (|g| \,\P |g|)
\le \inorm{k} \int b'_a \,d\P \, \int \sup_{g\in\fG_2} |g| \,d\P^*
\le  \inorm{k}^2 \bigl(\int b'_a \,d\P \bigr)^2 < \infty.
\end{eqnarray*}
Therefore, the sum on the right hand side in {(\ref{tmptmp3})} is finite and thus
the assumption $\P^* \snorm{g-\P g}^2_\fG < \infty$ is satisfied. This yields by part \emph{(iii)} of Theorem \ref{vdVW1996.Thm3.6.2} that
$\sup_{h\in \BL_1} \bigl| \Ex_M h(\Gnhat) - \Ex h(\bbG)\bigr|$ converges outer almost surely to zero and the sequence 
\be \label{tmptmp4}
 \Ex_M h(\Gnhat)^* - \Ex_M h(\Gnhat)_* 
\ee
converges almost surely to zero for every $h\in \BL_1$, where the asterisks denote the measurable cover functions with respect to $M$ and $Z_1, Z_2, \ldots$ jointly.\\

\emph{Step 4.~} 
Due to Step 3, the assumption {(\ref{vdVW3.9.9a})} of Theorem \ref{vdVW1996.Thm3.9.11} is satisfied.
We now show that additionally {(\ref{vdVW3.9.9b})} is satisfied,
i.e., that the term in {(\ref{tmptmp4})} converges to zero in outer probability.
In general, one can \emph{not} conclude that almost sure convergence implies convergence in outer probability, see \cite[p.\,52]{VandervaartWeller1996}. 
%
%
We know that the term in {(\ref{tmptmp4})} converges almost surely to zero for every $h\in \BL_1$, where the asterisks denote the \emph{measurable} cover 
functions with respect to $M$ and $(X_1,Y_1), (X_2,Y_2), \ldots$ 
\emph{jointly}. Hence, for every $h\in\BL_1$, the cover functions 
to be considered in {(\ref{tmptmp4})} are measurable.
Additionally, the multinomially distributed random variable $M$ 
is stochastically independent of $(X_1,Y_1), \ldots, (X_n,Y_n)$ 
in the bootstrap, where independence is understood in terms of a 
product probability space, see \cite[p.\,346]{VandervaartWeller1996} for details.
Therefore, an application of the Fubini-Tonelli theorem, see e.g., 
\cite[p.\,174, Thm.\,2.4.10]{DenkowskiEtAl2003}, yields that the inner integral
$\Ex_M h\bigl(\sqrt{n}(\Pnhat - \Pn)\bigr)^* 
 - \Ex_M h\bigl(\sqrt{n}(\Pnhat - \Pn)\bigr)_*$ 
considered by Fubini-Tonelli is \emph{measurable}
for every $n\in\N$ and every $h\in\BL_1$.
Recall that almost sure convergence of measurable functions implies 
convergence in probability which is equivalent with convergence in outer probability for measurable functions.
Hence we have convergence in outer probability in {(\ref{tmptmp4})}.
Therefore, all assumptions of Theorem \ref{vdVW1996.Thm3.9.11} are satisfied 
and the assertion of our theorem follows.  \hfill $\blacksquare$


\bibliographystyle{plain}

\end{document}